\pdfoutput=1

\documentclass[11pt]{article}
\usepackage{amsmath}
\usepackage{algorithmic}
\usepackage{algorithm}
\usepackage{multicol}
\usepackage[preprint]{acl}
\usepackage{enumitem}

\usepackage{times}
\usepackage{latexsym}

\usepackage[T1]{fontenc}

\usepackage[utf8]{inputenc}

\usepackage{microtype}

\usepackage{inconsolata}

\usepackage{graphicx}

\newcommand{\ie}{\emph{i.e.}}

\title{Paper Copilot: A Self-Evolving and Efficient LLM System for Personalized Academic Assistance}

\author{
 \textbf{Guanyu Lin\textsuperscript{1 2*}},
 \textbf{Tao Feng\textsuperscript{1*}},
 \textbf{Pengrui Han\textsuperscript{1 3*}},
 \textbf{Ge Liu\textsuperscript{1}},
 \textbf{Jiaxuan You\textsuperscript{1}}
\\
 \textsuperscript{1}University of Illinois at Urbana-Champaign,
 \textsuperscript{2}Carnegie Mellon University,
 \textsuperscript{3}Carleton College
 \\
 \textsuperscript{*}Equal contribution, Work done as intern
}

\begin{document}
\maketitle
\begin{abstract}
As scientific research proliferates, researchers face the daunting task of navigating and reading vast amounts of literature. Existing solutions, such as document QA, fail to provide personalized and up-to-date information efficiently. We present Paper Copilot, a self-evolving, efficient LLM system designed to assist researchers, based on thought-retrieval, user profile and high performance optimization. Specifically, Paper Copilot can offer personalized research services, maintaining a real-time updated database. Quantitative evaluation demonstrates that Paper Copilot saves 69.92\% of time after efficient deployment. This paper details the design and implementation of Paper Copilot, highlighting its contributions to personalized academic support and its potential to streamline the research process. We have deployed Paper Copilot at: \url{https://huggingface.co/spaces/ulab-ai/ArxivCopilot}.
\end{abstract}

\section{Introduction}

As scientific research has proliferated at an unprecedented rate, researchers are now supposed to navigate and interpret vast amounts of published and pre-print papers~\citep{tenopir2009electronic}. Indeed, researchers need to keep up with the latest trend. This involves continuously searching for relevant papers, quickly evaluating which papers for thorough reading, analyzing trending research topics, and reflecting potential ideas. Therefore, they should dedicate significant time to following up the latest papers. However, the large volume of papers make it hard for them to locate the related information, resulting in the waste of time. 

Fortunately, based on retrieval-augmented generation (RAG)~\citep{weijia2023replug}, LLMs~\citep{zhao2023survey} can help to extract and summarize useful information from such external papers~\citep{chen2023walking}. Thus, the above background leads us to a crucial question: \textit{How can we design a LLM system that can assist researchers in obtaining the latest research information from massive papers?}

\begin{figure}[t]
    \centering
    \includegraphics[width=\linewidth]{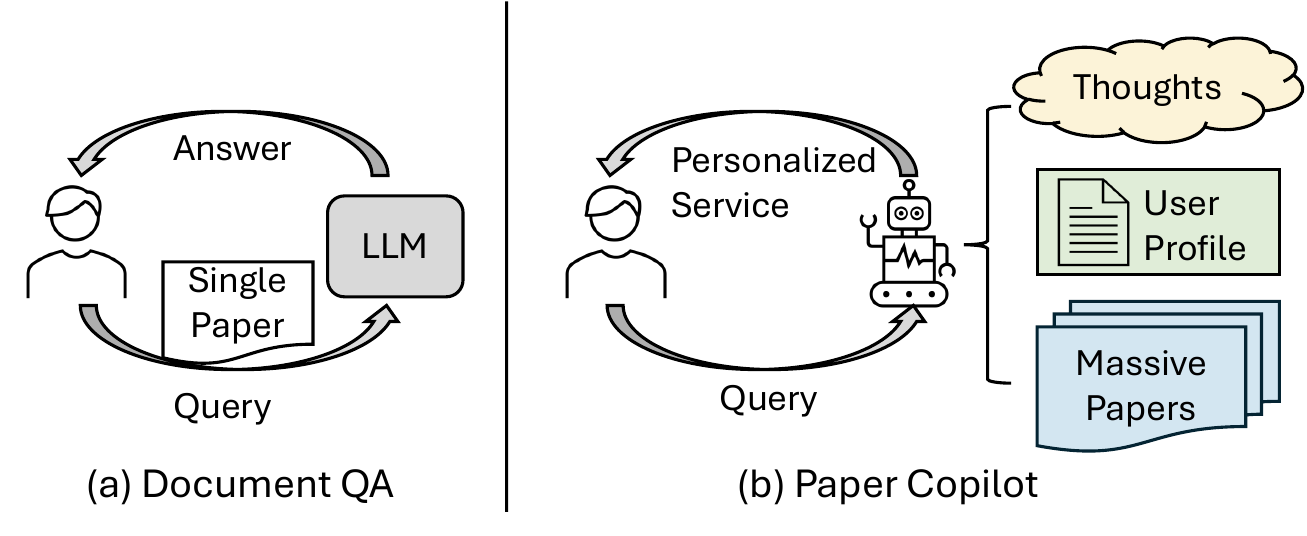}
    \caption{\textbf{Comparison of (a) document Question Answering (QA) with our (b) Paper Copilot.} Conventional document QA tends to help user understand the content of specific paper while our Paper Copilot can further act like a real research assistant who can provide personalized service based on user profile.}
\label{fig:compare_qa_ac}
\vspace{-0.5cm}
\end{figure}

To provide intelligent assistance for researchers, existing works have targeted several tasks, such as skimming~\citep{fok2023scim}, searching~\citep{ammar2018construction, beel2009google}, and reading~\citep{head2021augmenting}. However, these approaches focus either on understanding the content of paper document (as shown in Figure~\ref{fig:compare_qa_ac} (a)) or improving the ranking of relevant papers. They fall short of acting like a real researcher who can get \textit{personalized} and \textit{up-to-date} information on demand. Moreover, as researchers read more papers, they become increasingly experienced—a characteristic that current systems fail to replicate through \textit{self-evolution}. Finally, \textit{efficiency} remains a critical challenge in retrieving and extracting useful information from the vast and continuously growing pool of papers. 

To address the above challenges, we develop Paper Copilot, a self-evolving and efficient LLM system for personalized academic assistance. More specifically, Paper Copilot can provide personalized research service, self-evolve like a human researcher as shown in Figure~\ref{fig:compare_qa_ac} (b), and make prompt responses. The detailed characteristics of Paper Copilot are as below. 
\begin{itemize}[leftmargin=*]
    \item \textbf{Personalized research service}. Paper Copilot can provide personalized research assistance based on user profile. Specifically, it can (1) derive your profile from your historical publications, (2) analyze the latest trending research topics and provide ideas (which will be sent with email if sign up), and (3) offer research chat and advisory services.
        \item \textbf{Real-time updated research database}. Paper Copilot could refresh its paper database daily from the latest Arxiv papers. Users further have the option to select a date range to query the papers.

    \item \textbf{Self-evolved thought retrieval}. Paper Copilot enhances the response of LLM based on a thought retrieval~\citep{feng2024thought} method, which will self-evolve based on the historical user query.
    \item \textbf{High performance optimization}. Paper Copilot employs a real-time feature pool for efficient retrieval, a multithreading engine for effective memory management and I/O, and a cache to store responses with a high probability of re-querying. These optimizations significantly reduce API cost and response time by 69.92\%.
\end{itemize}
More importantly, user comment feedback indicates that Paper Copilot can save researchers at least 20 minutes in obtaining the same amount of information. This demonstrates that Paper Copilot not only provides valuable academic assistance but also saves researchers' time. Our evaluations, both quantitative and qualitative, further highlight its superiority in efficiency and user experience. Specifically, we reduce 69.92\% of time cost after efficient deployment. In summary, this work presents the following \textit{contributions}:
\begin{itemize}[leftmargin=*]
\item We design Paper Copilot, a self-evolving demo that provides personalized academic services based on real-time updated Arxiv papers.
\item We improve the efficiency and scalability of Paper Copilot through retrieval feature pre-computation, parallel computation, asynchronous I/O, and frequent query caching.
\item We evaluate the proposed Paper Copilot from both qualitative and quantitative perspectives.
\end{itemize}


\section{Paper Copilot}
\begin{figure}[t]
    \centering
    \includegraphics[width=1\linewidth]{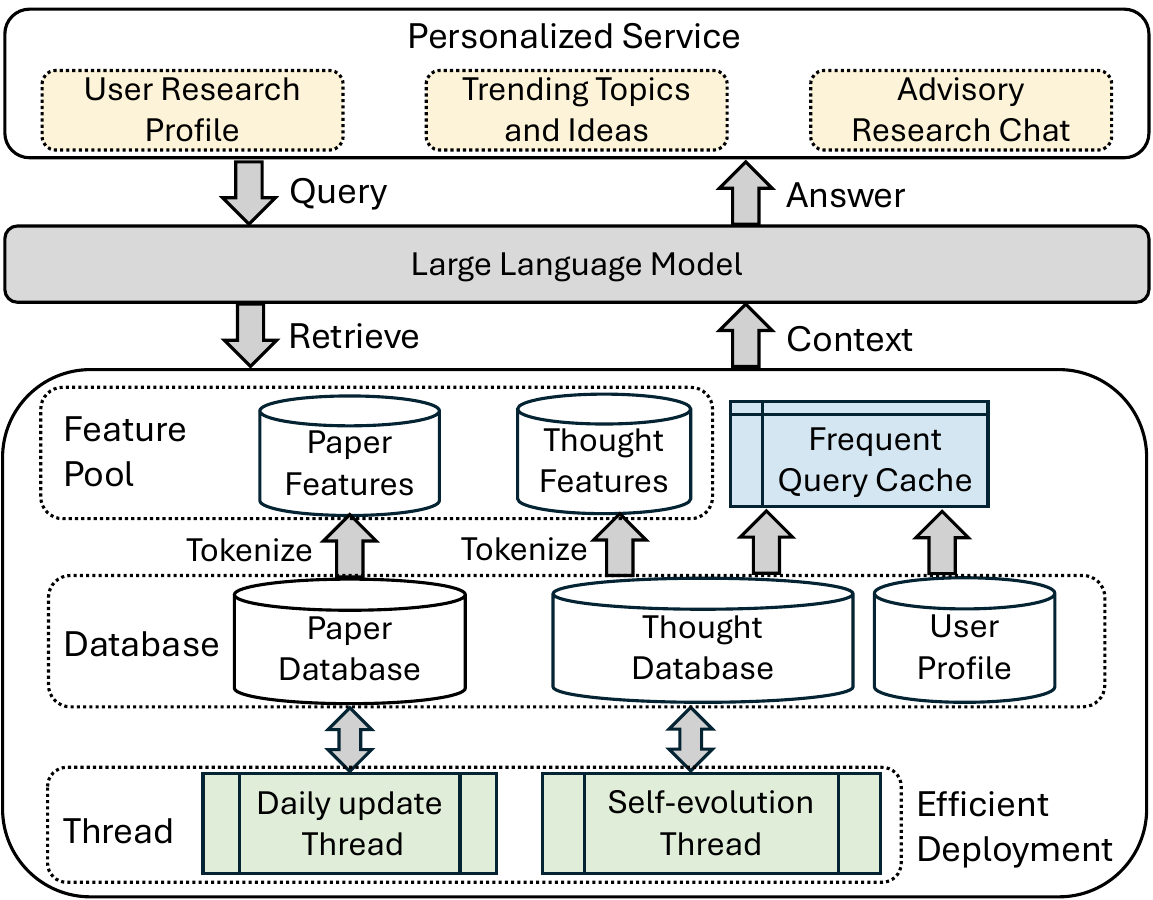}
    \caption{\textbf{Architecture of Paper Copilot from bottom-to-up perspective.} (a) In personalized service, Paper Copilot provides interactive services including the generation of user research profile, analysis of research trends and ideas, and advisory chatting about research. (b) In large language model, user demand from interaction will be used for retrieving and collecting relevant context, and then LLM will generate answer and make response to user demand. (c) In efficient deployment, feature pre-computation, parallel computation and caching techniques are applied to speed up the retrieval process and guarantee the efficient response.}
\label{fig:architecture}
\end{figure}
As shown in Figure~\ref{fig:architecture}, our proposed Paper Copilot mainly consists of the following four key parts:
\begin{itemize}
    \item \textbf{Personalized Service}. This part aims to generate personalized response based on user demand, including the generation of user research profile, analysis of personalized trending research topics or ideas with email, and personalized chat about research advisory.
    \item \textbf{Real-time Updating}. This part allows for the daily updating of its database using the latest Arxiv papers. Additionally, users can specify a range of time for papers to be retrieved.

    \item \textbf{Self-evolution}. This part improves LLM responses using a thought retrieval technique that adapts and evolves from past user queries.
        \item \textbf{Efficient Deployment}. This part achieves efficient deployment by a constantly updating feature pre-computation node for swift retrieval, a high performance engine for memory and I/O management, and a cache for storing frequently queried responses. 
\end{itemize}
For the detailed description of them, we will introduce in the subsequent section.
\subsection{Personalized Service}
\paragraph{User Research Profile} In user research profile, each user $u \in \mathcal{U}$ can input his/her name $n_u$ to get historical publication as:
$\mathcal{D}_{u, :t -1} \gets {\textbf{Search}}\left(n_u\right)$. Here ${\textbf{Search}()}$ is the search method based on Arxiv API~\cite{}. The retrieved papers $\mathcal{D}_{u, :t -1}$ will then be fed into LLM for profile generation as below.
\begin{equation}
    \mathcal{P}_{u, t} \gets {\textbf{LLM}}\left({\rm Instruct}_{p}, \mathcal{D}_{u, : t - 1}\right).
\end{equation}
where $\mathcal{P}_{u, t}$ is the generated profile for user $u$ at time step $t$. Besides, ${\rm Instruct}_{p}$ is the instruction for profile generation, which is defined in Section~\ref{tab:instruct_profile}.
\paragraph{Trending Topics and Ideas}
To further get the personalized trending research topics based on user profile, we firstly can retrieve some papers related to user profile $\mathcal{P}_{u, t}$, as follows: 
\begin{equation}\label{eq:re_trend}
\small
    \mathcal{R}^{trend}_{u, t} \gets {\textbf{Rtri}}\left(\textbf{Tkn}\left(\mathcal{P}_{u, t}\right), \textbf{Tkn}\left(\mathcal{D}_{:, : t - 1}\right)\right),
\end{equation}
where $\mathcal{R}^{trend}_{u, t}$ are the retrieved papers related to user profile. Besides, ${\textbf{Rtri}}()$ and $\textbf{Tkn}()$ are the methods for retrieval and tokenization. Based on the retrieved papers $\mathcal{R}^{trend}_{u, t}$, we can then feed them into LLM to generate the personalized trending research topics as below.
\begin{equation}
 \mathcal{C}_{u, t} \gets {\textbf{LLM}}\left({\rm Instruct}_{t}, \mathcal{R}^{trend}_{u, t}\right)   
\end{equation}
where $\mathcal{C}_{u, t}$ are the personalized trending research topics and ${\rm Instruct}_{t}$ is the instruction for research topic generation defined at Section~\ref{tab:instruct_trend}. With the personalized trending research topics, we can finally get some ideas related to the research topics of user $u$, as:
\begin{equation}
\mathcal{I}_{u, t} \gets {\textbf{LLM}}\left({\rm Instruct}_{i}, \mathcal{C}_{u, t}\right),
\end{equation}
where $\mathcal{I}_{u, t}$ are the research ideas related to the personalized trending research topics $\mathcal{C}_{u, t}$ of user $u$. Here ${\rm Instruct}_{i}$ is the instruction for idea generation defined at Section~\ref{tab:instruct_idea}. Besides, we also provide weekly report service for trending topics and ideas if users sign up with email.
\paragraph{Advisory Research Chat} In advisory research chat, user can further input his/her question $\mathcal{Q}_{u, t}$ and get personalized assistance based on previous generated trends and ideas. Firstly, we need to retrieve historical papers and generated contents $\mathcal{R}^{chat}_{u, t}$ related to the input question as:
\begin{equation}\label{eq:re_chat}
\small
    \mathcal{R}^{chat}_{u, t} \gets \textbf{Rtri}(\textbf{Tkn}\left(\mathcal{Q}_{u, t}\right), [\textbf{Tkn}\left(\mathcal{D}_{:, : t - 1}\right),  \textbf{Tkn}\left(\mathcal{B}_{:, :t - 1}\right)]),
\end{equation}
where $\mathcal{B}_{:, :t - 1} =  \mathcal{C}_{:, :t - 1}\cup \mathcal{I}_{:, :t - 1}\cup \mathcal{A}_{:, :t - 1}$ is the thought database including generated research trends $\mathcal{C}_{:, :t - 1}$, ideas $\mathcal{I}_{:, :t - 1}$, and answers $\mathcal{A}_{:, :t - 1}$. Based on the retrieved historical papers and generated contents, we can then feed them into LLM for answering:
 \begin{equation}
     \mathcal{A}_{u, t} \gets  {\textbf{LLM}}\left(\mathcal{Q}_{u, t}, \mathcal{R}^{chat}_{u, t}, \mathcal{P}_{u, t}\right)
 \end{equation}
 where $\mathcal{A}_{u, t}$ is the answer for user $u$ based on his/her question $\mathcal{Q}_{u, t}$. Here feeding $\mathcal{P}_{u, t}$ into LLM means the generated answer will be organized in a personalized manner related to the profile of user $u$.
\subsection{Real-time Updating}

\paragraph{Daily Updating} During daily updating, Paper Copilot will download the newest papers from Arxiv and refresh the paper storage as:
$\mathcal{D}_{: , :t} \gets \mathcal{D}_{:, :t - 1} \cup \mathcal{D}_{: , t}$, where $\mathcal{D}_{: , t}$ are the newest papers and $\mathcal{D}_{: , :t}$ is the refreshed paper storage.

\paragraph{Time Range Selection} As users may not care about some old papers and trends. Thus, in time range selection, users can select the daily papers $\mathcal{D}_{:, t}$, weekly papers
$\mathcal{D}_{:, t - 6 : t}$, and all papers $\mathcal{D}_{:, :t}$ for personalized research trend and idea generation.

\subsection{Self-evolution} As human researchers will become more and more experienced, Paper Copilot also evolves its thought by incorporating the interacted contents with users as below.
\begin{equation}
    \begin{aligned}
        \mathcal{A}_{: , :t} & \gets \mathcal{A}_{:, :t - 1} \cup \mathcal{A}_{: , t}, \\ 
\mathcal{C}_{: , :t} &\gets \mathcal{C}_{:, :t - 1} \cup \mathcal{C}_{:, t}, \\
\mathcal{I}_{: , :t} &\gets \mathcal{I}_{:, :t - 1} \cup \mathcal{I}_{:, t},
    \end{aligned}
\end{equation}
where $\mathcal{A}_{: , :t}$, $\mathcal{C}_{: , :t}$, and $\mathcal{I}_{: , :t}$ are the self-evolved thought at time step $t$ by incorporating answers, research trends and ideas interacted with users. That is to say, the more interactions with users, the smarter Paper Copilot will be.

\subsection{Efficient Deployment}
\begin{figure}[!htb]
    \centering
    \includegraphics[width=.9\linewidth]{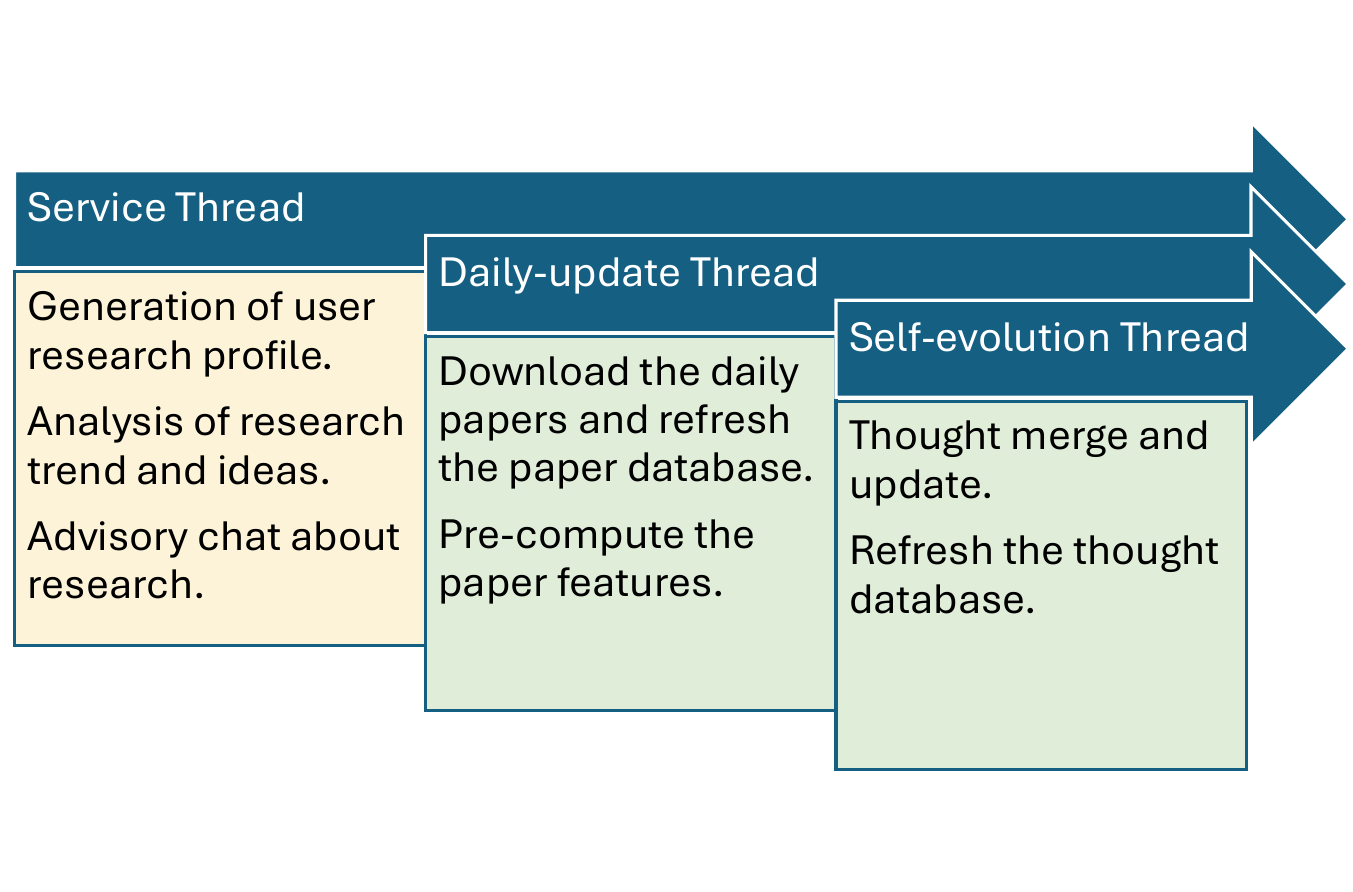}
    \caption{\textbf{Multi-thread engine keeps Paper Copilot service away from waiting for daily updating of papers and self-evolution of thoughts.} The daily-update thread and self-evolution thread will achieve thought memory management and asynchronous I/O without disturbing the service thread.}
\label{fig:mul_thread}
\end{figure}
\paragraph{Feature Pre-computation}
In feature pre-computation, we construct a feature pool and pre-compute the paper embedding $\mathbf{D}_{:, : t - 1}$ and thought embedding $\mathbf{B}_{: , :t - 1}$ for retrieval. By this way, we do not need to re-tokenize the input text while retrieval, which saves a lot of time. Thus the retrieval equations at Eq.~\eqref{eq:re_trend} and \eqref{eq:re_chat}, respectively, can be reformulated as Eq.~\eqref{eq:re_for_trend} and \eqref{eq:re_for_idea}.
\begin{equation}\label{eq:re_for_trend}
\small
        \mathcal{R}^{trend}_{u, t} \gets {\textbf{Rtri}}\left(\textbf{Tkn}\left(\mathcal{P}_{u, t}\right), \mathbf{D}_{:, : t - 1}\right), 
\end{equation}
\begin{equation}\label{eq:re_for_idea}
\small
    \mathcal{R}^{chat}_{u, t} \gets \textbf{Rtri}\left(\textbf{Tkn}\left(\mathcal{Q}_{u, t}\right), [\mathbf{D}_{:, : t - 1}, \mathbf{B}_{: , :t - 1}]\right),
\end{equation}
where the computational costs for the tokenization methods on papers $\mathcal{D}_{:, : t - 1}$ and thought $\mathcal{B}_{:, :t - 1}$ are saved. Besides, the paper embedding and thought embedding will be updated through:
\begin{equation}
    \mathbf{D}_{: , :t} \gets [\mathbf{D}_{:, : t - 1}, \textbf{Tkn}\left(\mathcal{D}_{: , t}\right)],
\end{equation}
\begin{equation}
    \begin{aligned}
       \mathbf{A}_{: , :t} \gets [\mathbf{A}_{:, : t - 1}, \textbf{Tkn}\left(\mathcal{A}_{: , t}\right)], \\
\mathbf{C}_{u , :t} \gets [\mathbf{C}_{u, : t - 1}, \textbf{Tkn}\left(\mathcal{C}_{: , t}\right)], \\
\mathbf{I}_{u , :t} \gets [\mathbf{I}_{u, : t - 1}, \textbf{Tkn}\left(\mathcal{I}_{: , t}\right)], \\
\mathbf{B}_{: , :t} \gets [\mathbf{A}_{: , :t}, \mathbf{C}_{: , :t}, \mathbf{I}_{: , :t}],
    \end{aligned}
\end{equation}
where $\mathbf{D}_{: , :t}$ and $\mathbf{B}_{: , :t}$ are the updated paper embedding and thought embedding, respectively.

\paragraph{Multi-threading Engine}

As our Paper Copilot needs to refresh the database and update thoughts frequently, the user interactive service will be disturbed and become inefficient. Thus we further implement a multi-thread engine as Figure~\ref{fig:mul_thread} to reduce the waiting time of interactive service when updating. Specifically, it consists of service thread, daily-update thread and self-evolution thread to execute the personalized service, paper updating and thought management at the same time. With such multi-thread engine, there is no need for the main personalized service to wait for storage refreshing. That is to say, all memory management processes and I/O processes will be finished in parallel.

\paragraph{Frequent Query Cache} In frequent query cache, we store the content that will be frequently queried at hash cache. More specifically, user profile, research trends and ideas may will stay unchanged within a period of time. Thus these static contents are more likely to be re-queried, and we store them in hash cache ${\textbf{Hash}}()$ as:
\begin{equation}
\small
    \begin{aligned}
        \mathcal{P}_{u, t} \gets {\textbf{Hash}}\left(n_u\right), & 
        \mathcal{C}_{u, t} \gets {\textbf{Hash}}\left(\mathcal{P}_{u, t}\right), \\
        \mathcal{I}_{u, t} \gets {\textbf{Hash}}\left(\mathcal{P}_{u, t}\right), &
        \mathcal{R}^{trend}_{u, t} \gets {\textbf{Hash}}\left(\mathcal{P}_{u, t}\right),
    \end{aligned}
\end{equation}
where $\mathcal{R}^{trend}_{u, t}$ are the papers we retrieve for research trend generation. As $\mathcal{R}^{trend}_{u, t}$ will also be presented at Paper Copilot as trending papers, we hash them in the cache. With this hash cache, we can make instant responses when contents are re-queried.

\section{User Guidance and Usage}
\begin{figure}[!htb]
    \centering
\includegraphics[width=.95\linewidth]{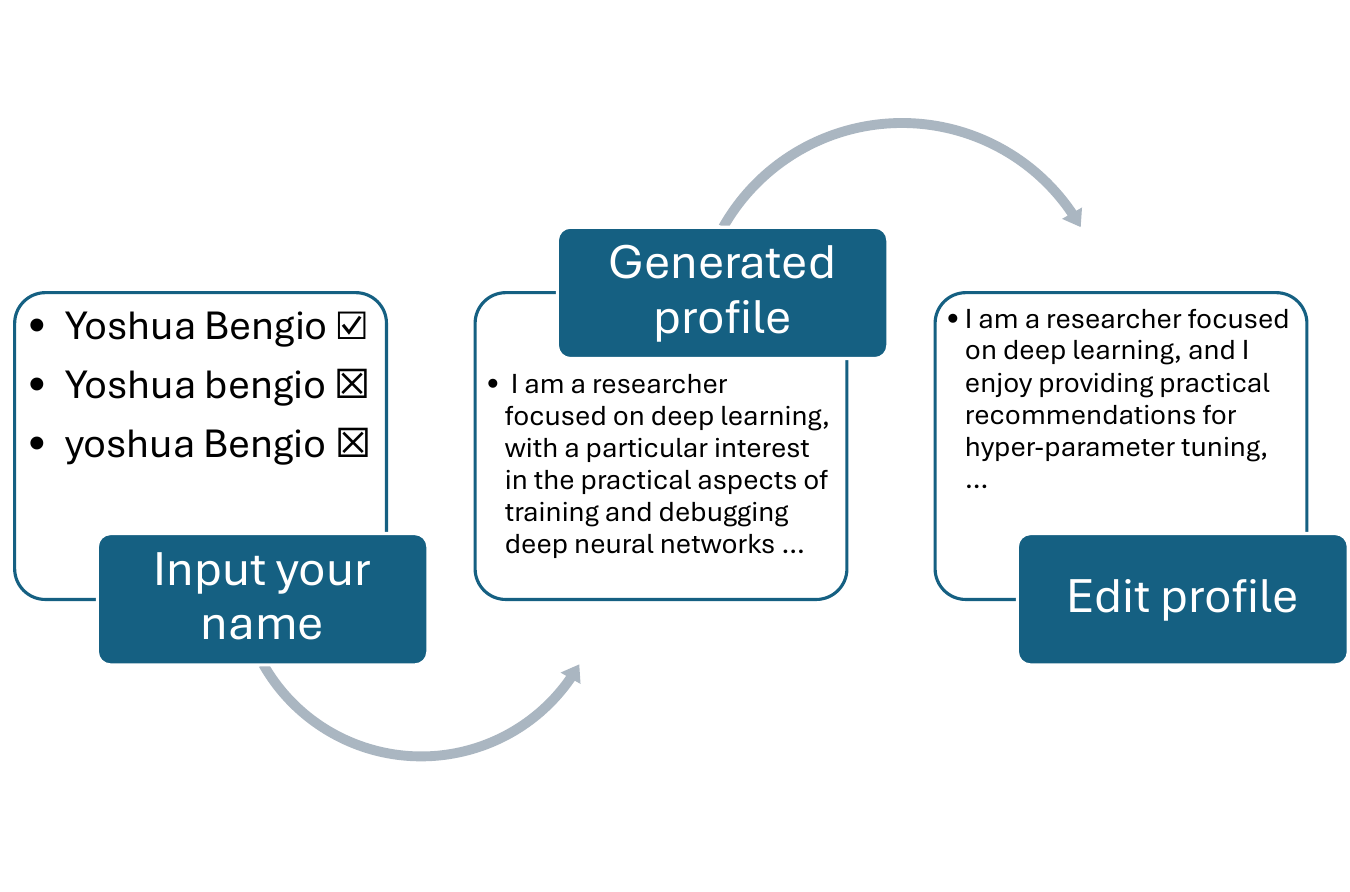}
    \caption{\textbf{Flowchart for the interaction of user research profile in Paper Copilot}. Users can input his/her name to generate the personalized profile based on historical publication. Besides, if users are unsatisfied with the generated profile or fail to get historical publication, they also can manually edit the profile.}
\label{fig:profile}
\end{figure}
\begin{figure*}[!htb]
    \centering
    \begin{tabular}{c|c}
           \includegraphics[width=.45\linewidth]{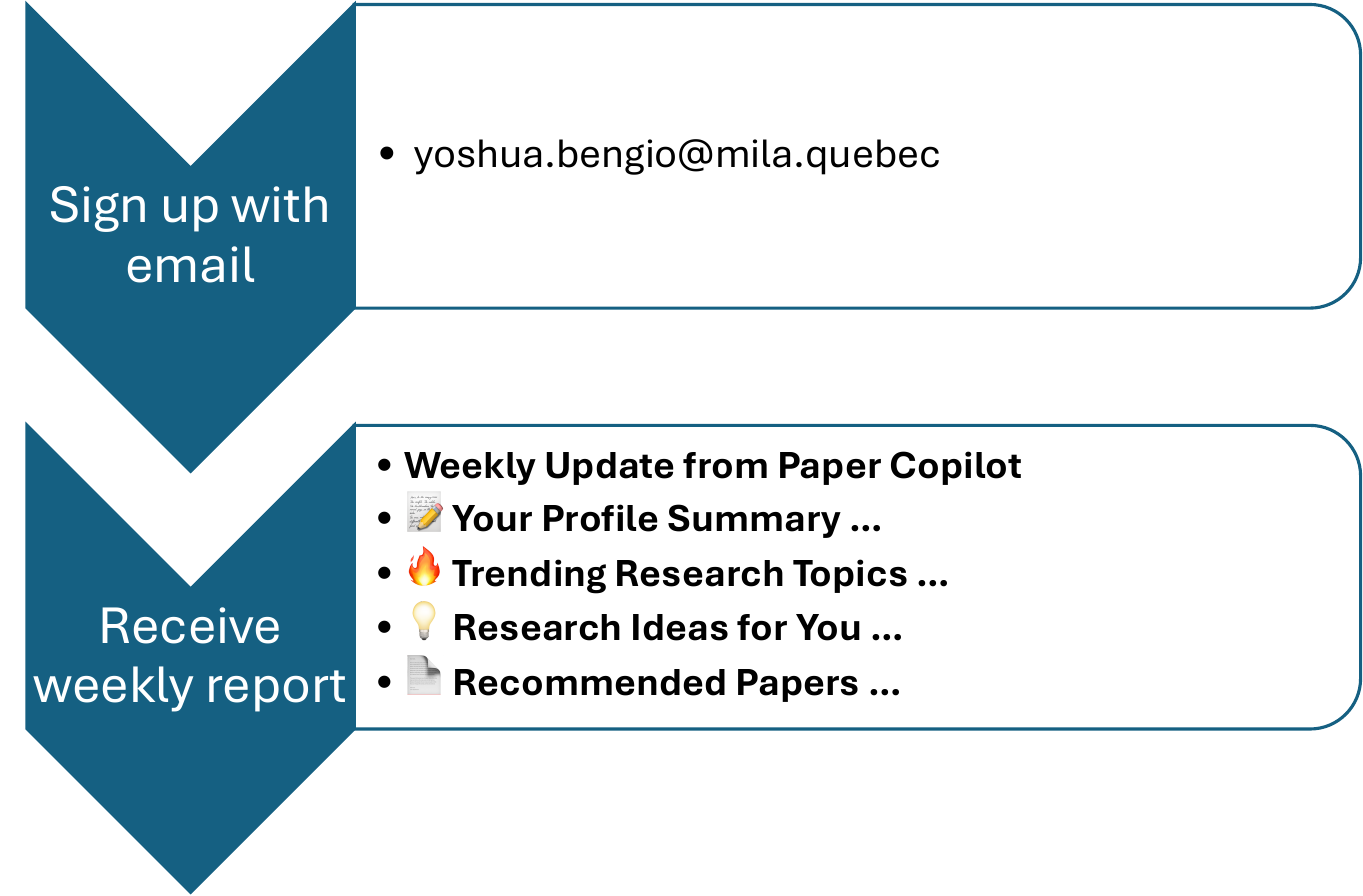}  & \includegraphics[width=.4\linewidth]{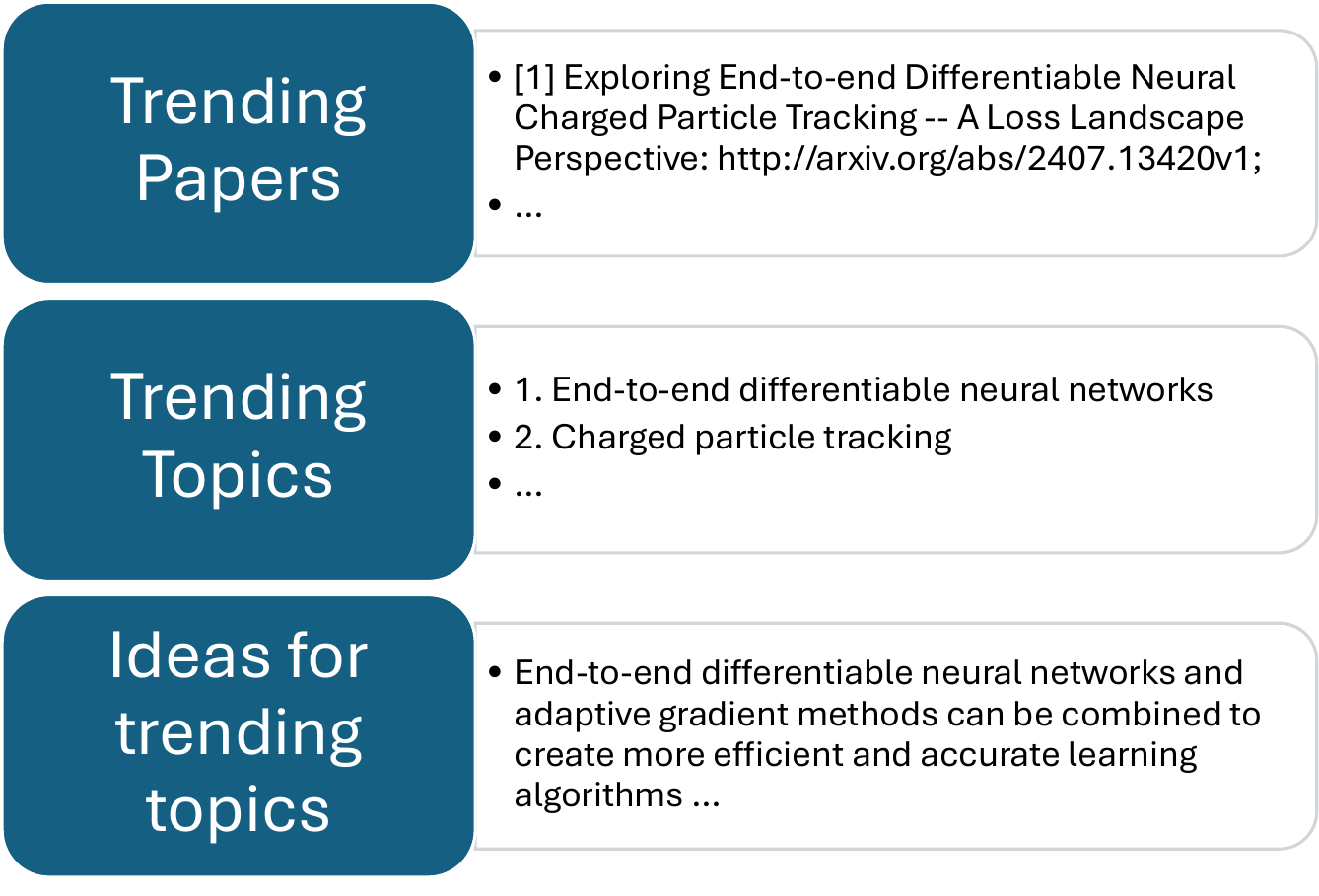} \\ 
         (a) Sign up with email & (b) Get research trend
    \end{tabular}

    \caption{\textbf{Diagram for the interaction of research trend and ideas in Paper Copilot}. (a) Users can sign up with email to receive the weekly update. (b) Besides, users can also select the time range for getting the daily, weekly or all historical research trend.}
\label{fig:trend}
\end{figure*}
\begin{figure}[!htb]
    \centering
    \includegraphics[width=.9\linewidth]{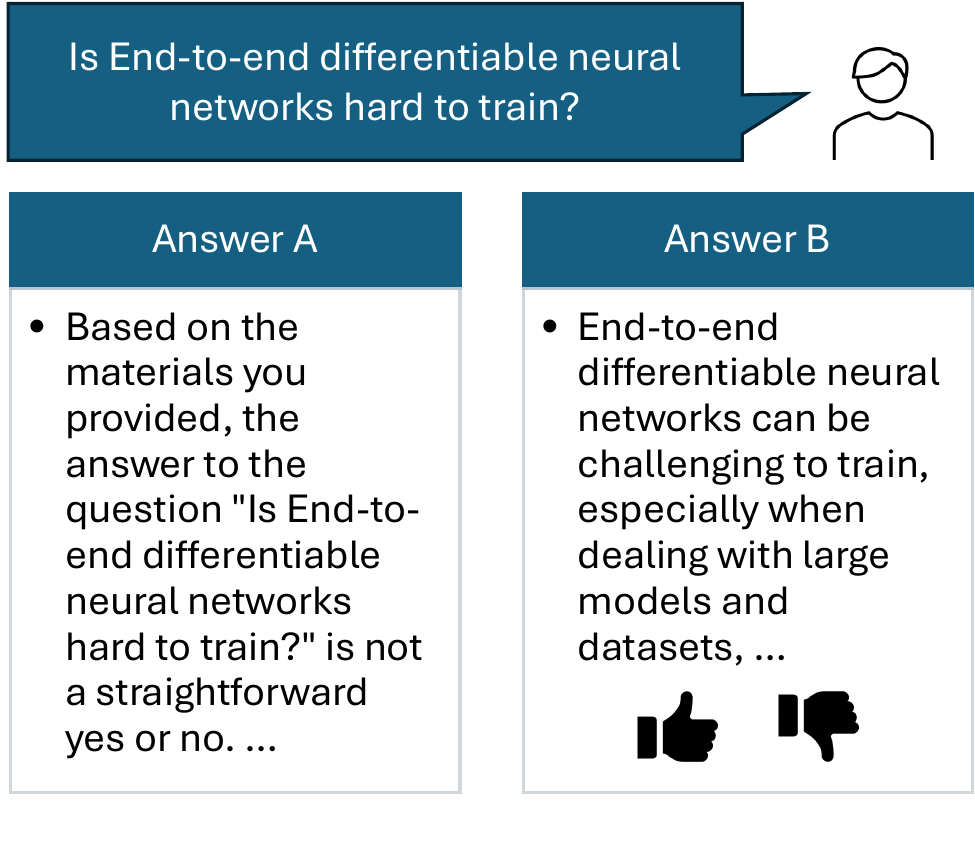} \\
    \includegraphics[width=.7\linewidth]{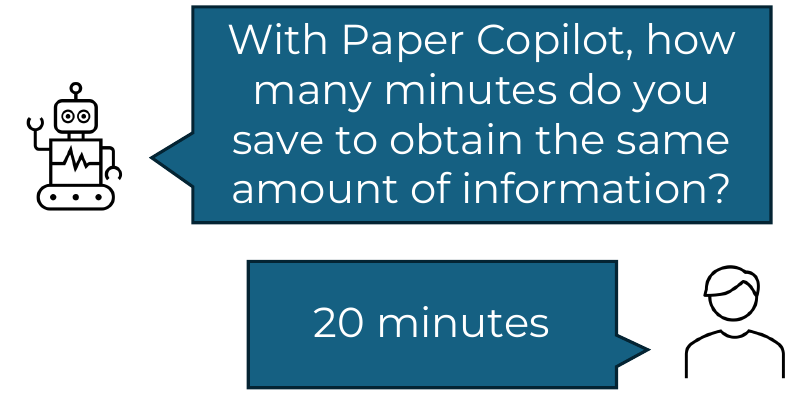}
    \caption{\textbf{Diagram for the interaction of advisory research chat in Paper Copilot}. After users ask the question, Paper Copilot will give two answers. Specifically, the first answer is with both thought and paper retrieval while the second answer is just with paper retrieval. Here the second answer will have two feedback choices for users, one is 'like' and another is 'dislike'. If users click 'like', the first answer will be removed. Otherwise, the second answer will removed. Besides, users can also provide feedback on the saved time.}
\label{fig:chat}
\end{figure}
\paragraph{\textbf{User Research Profile}} 
In "Set your profile!", as shown in Figure~\ref{fig:profile}, we have input text box "Input your name:" where user can input his/her name and then click button "Set Profile" to obtain the profile from output text box "Generated profile (can be edited):". Here the output text box of generated profile also can be modified and edited by clicking button "Edit Profile". The details of each button operation is shown in Figure~\ref{fig:scrn_profile} of Appendix~\ref{sec:appendix}.

\paragraph{Trending Topics and Ideas} In "Get trending topics and ideas!", as shown in Figure~\ref{fig:trend}, user can sigu up to get the weekly update of trending research topics, ideas and papers. Besides, user can also select the time range and then click button "Confirm" to filter out papers from daily, weekly and all historical publication time. Then in the "Trending Papers", "Trending Topics" and "Ideas for Trending Topic" text boxes, respectively, personalized trending papers, topics and ideas related to the user will be presented. The details of each button operation is shown in Figure~\ref{fig:scrn_trend} of Appendix~\ref{sec:appendix}.

\paragraph{\textbf{Advisory Research Chat}} In "Chat with Paper Copilot!", as shown in Figure~\ref{fig:chat}, user can chat with Paper Copilot by typing the question into the input text box of Chatbot and then click button "Send" or enter "carriage return" in the keyboard. Then Paper Copilot will return with two candidate answers, the first answer is based on thought and paper retrieval while the second answer is just based on paper retrieval. Here user can give feedback and choose the preferred answer with either augmented thoughts or just initial papers. Besides, by clicking the button "Clear", user can clean all historical chat with Paper Copilot. Finally, user can give further feedback about how many minutes Paper Copilot has helped you to save time in research by clicking button "Comment". The details of each button operation is shown in Figure~\ref{fig:scrn_chat} of Appendix~\ref{sec:appendix}.

\section{Evaluation}
\begin{figure}[!htb]
    \centering
    \includegraphics[width=.9\linewidth]{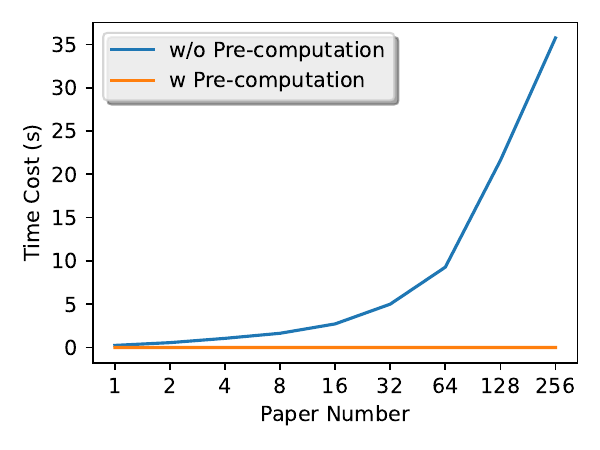}
    \caption{\textbf{Feature pre-computation significantly improves the efficiency.} The time cost for retrieval without feature pre-computation will grow with the exponential increase of paper number, while our proposed feature pre-computation stays unchanged and keeps constant time cost.}
\label{fig:time}
\end{figure}
\paragraph{\textbf{Quantitative: Efficiency}}
Firstly, as shown in Figure~\ref{fig:time}, we plot the time costs of paper retrieval without feature pre-computation and with pre-computation. From the result, we
can discover that our proposed feature pre-computation is very efficient, which has a constant computational cost at $O(1)$. However, the time cost of retrieval without pre-computation will grow significantly with the increase of papers. This is because there is no need to re-tokenization on contents to be retrieved under feature pre-computation, while those without pre-computation will repeatedly tokenize the contents each time.
\begin{figure}[!htb]
    \centering
    \includegraphics[width=.85\linewidth]{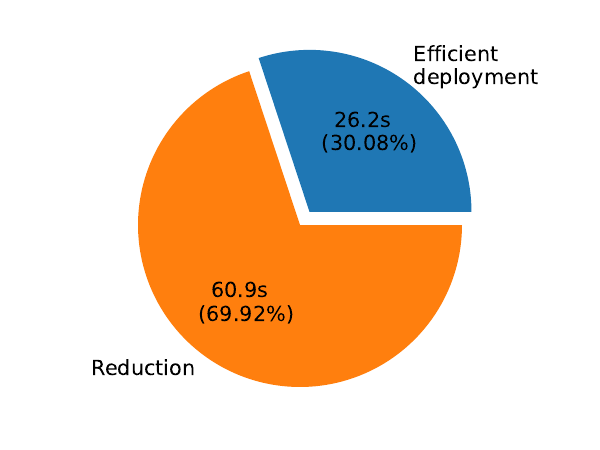}
    \caption{\textbf{Efficient deployment methods dramatically reduce the time cost.} The average total time cost before efficient deployment is 87.1s (26.2s + 60.9s), which is reduced by 69.92\% after efficient deployment.}
\label{fig:eff_time}
\end{figure}

Besides, we also plot the pie chart of time cost reduced by efficient deployment and that under efficient deployment as Figure~\ref{fig:eff_time}. Specifically, we can see that our efficient deployment reduces the total time cost average by 60.9s. And now is just requires average 26.2s for making response, which improves the user experience a lot compared with initial 87.1s.
\paragraph{\textbf{Qualitative: User Study}}
After collecting the user feedback from advisory research chat, we find that there are about 75\% of users will prefer the answers with self-evolution augmentation, illustrating the effectiveness of Paper Copilot for self-evolving like real human researchers.

However, there is still a small problem. That is, when user inputs his/her name in profile generation, there may be duplicate. For example, when you input "Feifei Li", you will get the profile of a researcher in quantum computing, instead of the researcher in artificial intelligence. In such case, the users may need to input and edit the profile manually by themselves.

\section{Related Work}
\paragraph{Retrieval Augmented Generation}
Retrieval Augmented Generation (RAG)~\cite{lewis2020retrieval} augments LLMs by retrieving and incorporating external context and information. Existing approaches employ methods can be classified into the following categories: embedding-based method \citep{contriever,dragon}, fine-tuning re-ranker method \citep{ram2023context} and keyword-based method \citep{robertson2009probabilistic}. While these strategies have shown decent outcomes, they still face many challenges in the extremely long context. Fortunately, hierarchical tree-based method \citep{chen2023walking} and thought-retrieval method \citep{feng2024thought} can well address these challenges. Though extending the long context window, existing method is still inefficient when encoding the extremely long context. Thus, in this work, we further improve the efficiency of long-context RAG by feature pre-computation and several high performance computing techniques.

\paragraph{Academic Assistance with Language Models}
Language models can provide academic assistance based on scientific papers in variety of ways. Firstly, it can make summary of the paper's content to help understanding~\citep{nenkova2012survey, sefid2022scibertsum}. Besides, it also can help researchers to skim today's emerging papers~\citep{fok2023scim} and read useful information~\citep{august2023paper}. However, existing works mainly focus on single paper understanding. Unlike them, Paper Copilot further provides personalized academic assistance like a human researcher.

\section{Conclusion and Future Work}
To address the challenges posed by the rapid growth of scientific research, we propose Paper Copilot with a personalized, self-evolving, and efficient LLM system. It offers tailored research services, maintains a real-time updated database, and employs advanced optimization techniques to enhance performance. Evaluations demonstrate its ability to significantly reduce the time researchers spend on literature review while improving accuracy and user experience. By setting a new standard for personalized academic support, Paper Copilot stands as a valuable tool for the scientific community, enhancing the research process. Future work will focus on integrating additional sources beyond Arxiv to provide a broader research perspective.


\bibliography{custom}
\clearpage
\appendix
\onecolumn

\section{Example Appendix}\label{sec:appendix}
\begin{table*}[!htb]
\centering
\caption{Prompts for profile generation.}
\label{tab:instruct_profile}
\begin{tabular}{c}
\hline
\begin{tabular}[c]{@{}l@{}}Instruction:  Based on the list of the researcher's papers from different periods, please write a \\comprehensive first person persona. Focus more on recent papers. Be concise and clear \\(around 300 words). \\ \\ Here are the papers from different periods: \{papers\}\end{tabular} \\ \hline
\end{tabular}
\end{table*}

\begin{table}[!htb]
\centering
\caption{Prompts for trending research topic generation.}
\label{tab:instruct_trend}
\begin{tabular}{c}
\hline
\begin{tabular}[c]{@{}l@{}}Instruction:  Given some recent paper titles and abstracts. Could you summarize no more than \\10 top keywords of high level research backgrounds and trends.\\ \\ Here are the retrieved paper abstracts: \{papers\}\end{tabular} \\ \hline
\end{tabular}
\end{table}

\begin{table}[!htb]
\centering
\caption{Prompts for research idea generation.}
\label{tab:instruct_idea}
\begin{tabular}{c}
\hline
\begin{tabular}[c]{@{}l@{}}Instruction: Here is a high-level summarized trend of a research field: \{trend\} \\ \\ How do you view this field? Do you have any novel ideas or insights?\\ Please give me 3 to 5 novel ideas and insights in bullet points. Each bullet points should be \\concise, containing 2 or 3 sentences.\end{tabular} \\ \hline
\end{tabular}
\end{table}

       \begin{figure}[!htb]
    \centering
\includegraphics[width=.9\linewidth]{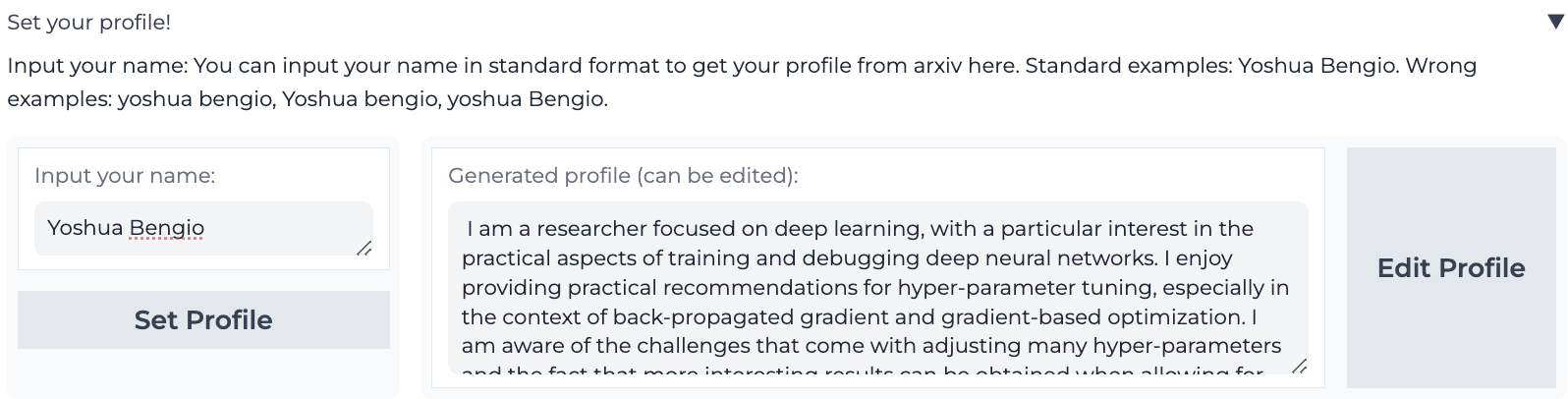}
    \caption{\textbf{Screenshot for the interaction of user research profile in Paper Copilot}. Users can input his/her name and then click "Set Profile" to generate the personalized profile based on historical publication. Besides, if users are unsatisfied with the generated profile or fail to get historical publication, they also can manually edit the profile and then click "Edit Profile".}
\label{fig:scrn_profile}
\end{figure}

\begin{figure}[!htb]
    \centering
\includegraphics[width=.8\linewidth]{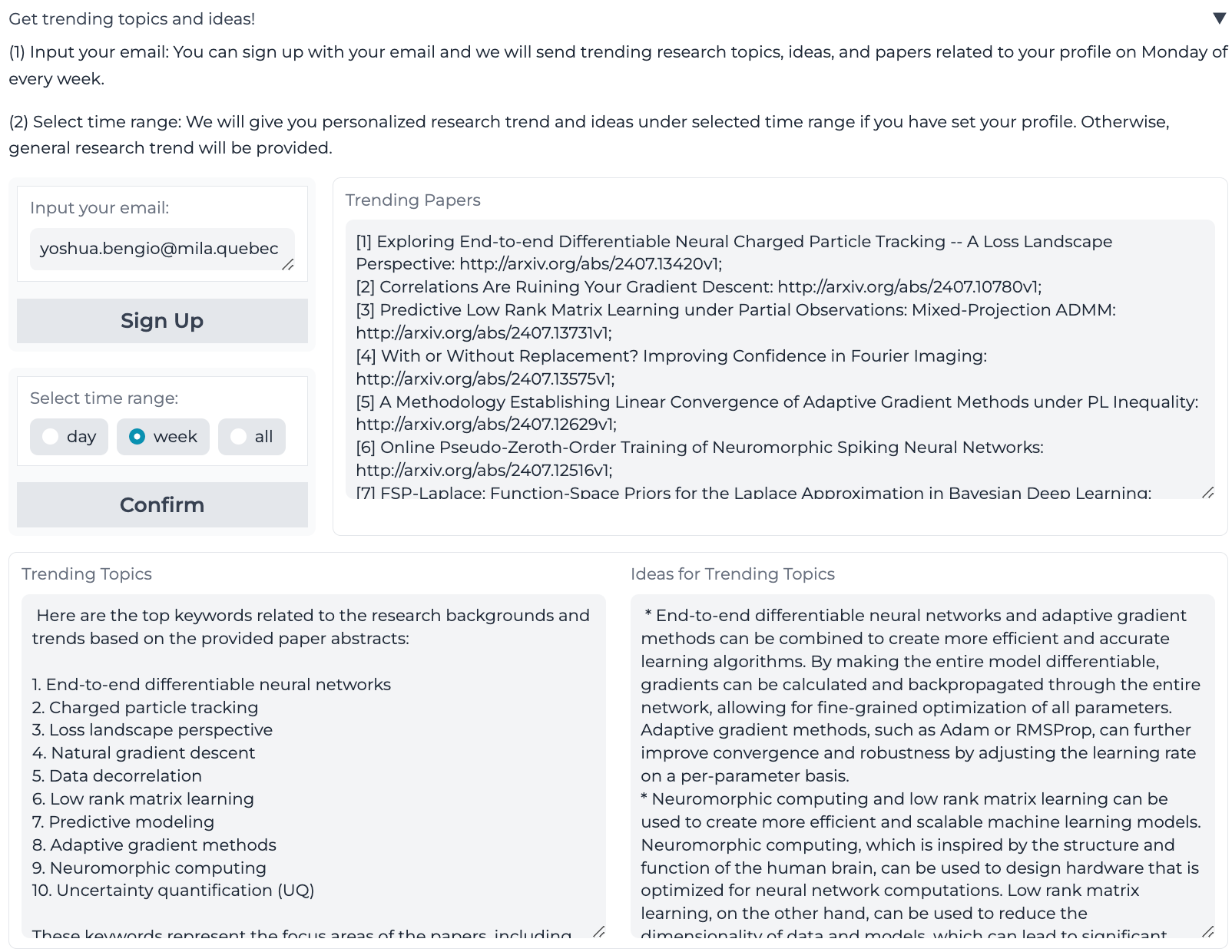}
    \caption{\textbf{Screenshot for the interaction of research trend and ideas in Paper Copilot}. Users can sign up with email to receive the weekly update. Besides, users can also select the time range for getting the research trend and we have three choices here \ie day means getting trend from today's papers, week means getting trend from this week's papers and all means getting trend from all papers. After selecting the time range, users can click "Confirm" and the trending papers, trending research topics and ideas will be shown to the users.}
\label{fig:scrn_trend}
\end{figure}

\begin{figure}[!htb]
    \centering
\includegraphics[width=.9\linewidth]{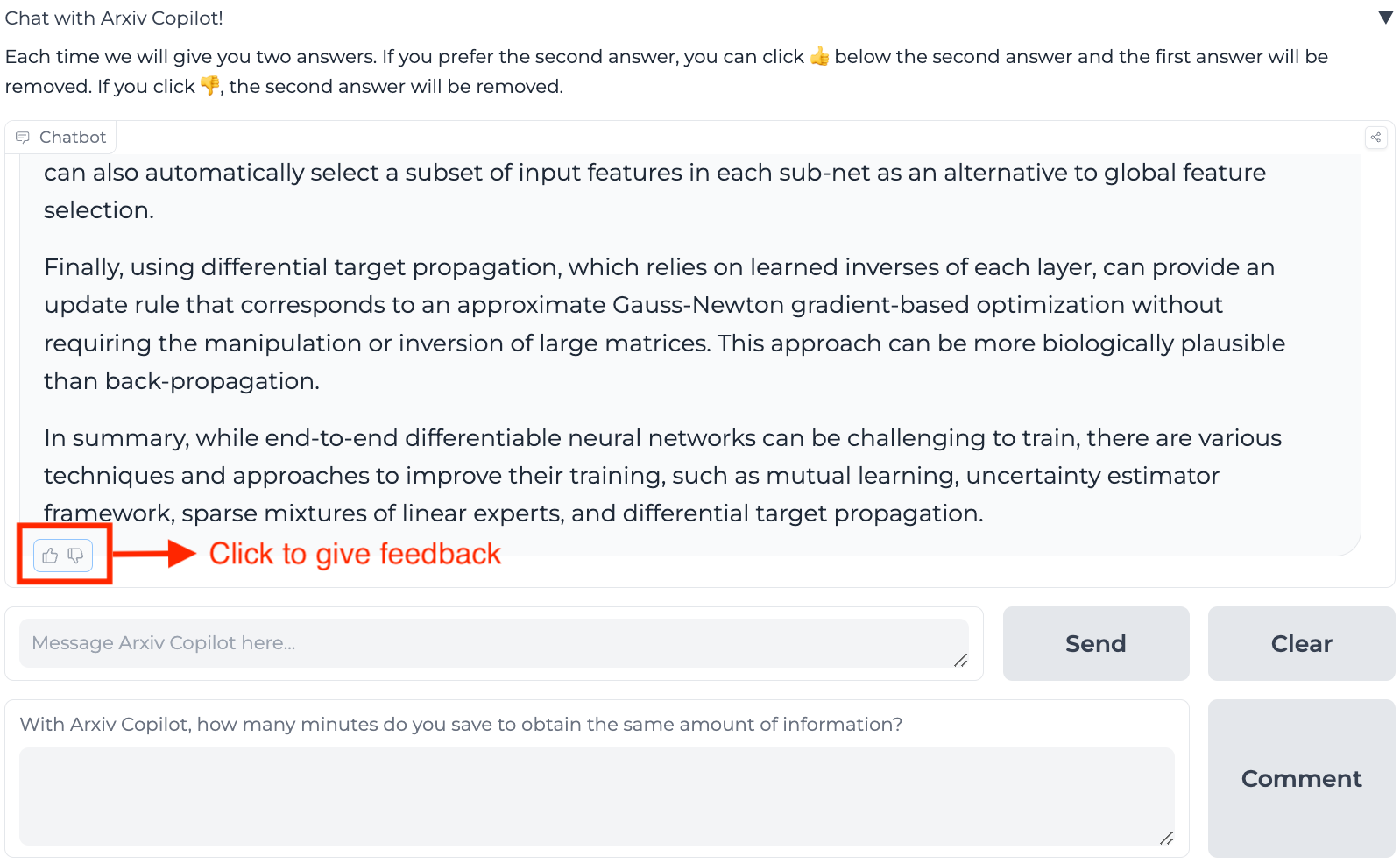}
    \caption{\textbf{Screenshot for the interaction of advisory research chat in Paper Copilot}. Users can click "send" after entering the question and Paper Copilot will give two answers. Specifically, the first answer is with both thought and paper retrieval while the second answer is just with paper retrieval. Here the second answer will have two feedback choices for users, one is 'like' and another is 'dislike'. If users click 'like', the first answer will be removed. Otherwise the second answer will removed. Besides, users can also clean the chat history by clicking "Clear" and provide further feedback by clicking "Comment".}
\label{fig:scrn_chat}
\end{figure}

\end{document}